\RequirePackage{fix-cm}
\documentclass[smallcondensed]{svjour3}       
\smartqed  
\usepackage[utf8]{inputenc} 
\usepackage[T1]{fontenc}    
\usepackage{hyperref}       
\usepackage{url}            
\usepackage{booktabs}       
\usepackage{amsfonts}       
\usepackage{nicefrac}       
\usepackage{microtype}      
\usepackage{xcolor}         
\usepackage[inline]{enumitem}
\usepackage{amsmath}
\usepackage{algorithm, algpseudocode}
\usepackage{amsmath,amsfonts,bm}

\newcommand{\D}{\mathcal{D}}

\newcommand{\E}{\mathbb{E}}

\newcommand{\calX}{\mathcal{X}}

\newcommand{\calY}{\mathcal{Y}}
\newcommand{\calL}{\mathcal{L}}
\newcommand{\calS}{\mathcal{S}}
\newcommand{\calT}{\mathcal{T}}

\newcommand{\calZ}{\mathcal{Z}}
\newcommand{\calN}{\mathcal{N}}

\newcommand{\calG}{\mathcal{G}}

\newcommand{\RN}[1]{%
  \textup{\uppercase\expandafter{\romannumeral#1}}%
}

\usepackage{caption}
\usepackage{wrapfig}
\usepackage{subcaption}
\usepackage{graphicx}

\begin{document}

\title{On the benefits of representation regularization in invariance based domain generalization}

\author{Changjian Shui, Boyu Wang, Christian Gagné}

\authorrunning{C.Shui, B.Wang, C.Gangé}

\institute{
        Changjian Shui 
        \at Université Laval
        \at  \email{changjian.shui.1@ulaval.ca}           
        \and
        Boyu Wang \at Western University, Vector Institute
        \at \email{bwang@csd.uwo.ca}
        \and
        Christian Gangé \at Université Laval, Canada CIFAR AI Chair, Mila
        \at  \email{christian.gagne@gel.ulaval.ca}  \\ \\
        Correspondence to Changjian Shui}

\date{Received: date / Accepted: date}

\maketitle
\begin{abstract}
A crucial aspect in reliable machine learning is to design a deployable system in generalizing new related but unobserved environments. Domain generalization aims to alleviate such a prediction gap between the observed and unseen environments. Previous approaches commonly incorporated learning invariant representation for achieving good empirical performance. In this paper, we reveal that merely learning invariant representation is vulnerable to the unseen environment. To this end, we derive novel theoretical analysis to control the unseen test environment error in the representation learning, which highlights the importance of controlling the smoothness of representation. In practice, our analysis further inspires an efficient regularization method to improve the robustness in domain generalization. Our regularization is orthogonal to and can be straightforwardly adopted in existing domain generalization algorithms for invariant representation learning. Empirical results show that our algorithm outperforms the base versions in various dataset and invariance criteria.
\keywords{Domain Generalization \and Transfer Learning \and Representation Learning}
\end{abstract}

\section{Introduction}
Most research in deep learning assumes that models are trained and tested from a fixed distribution. However, such deep models generally failed to adopt in the real-world applications, because the test environment is often different from training (or observed) environments. Thus, the capacity in generalizing the new environment is crucial for developing reliable and deployable deep learning algorithms (e.g \cite{goodfellow2014explaining}). 

To this end, \emph{Domain Generalization} is recently proposed and studied to alleviate the prediction gap between the observed training ($\calS$) and \emph{unseen} test ($\calT$) environments. Taking the advantage of the learned inductive bias from multiple observed sources, the prediction on the test environment can be guaranteed in some specific scenarios \cite{baxter2000model}. 

Meanwhile, extrapolation to a new environment is challenging since the environmental distribution-shifts are inevitable and unknown in advance. Such changes typically include covariate shift  \cite{sugiyama2007covariate}, conditional shift \cite{li2018domain,arjovsky2019invariant} or both. Based on different distribution-shift assumptions, a widely adopted principle is to learn a representation to satisfy several invariance criteria \cite{buhlmann2020invariance} among the observed environments (i.e, sources $\calS$). Through minimizing the source prediction risk and enforcing the invariance, the prediction performance can be improved in many empirical scenarios \cite{dg_mmld,li2018domain}.

Although learning invariance is popular in domain generalization with certain practical success, its theoretical counterpart still remains elusive. For instance, \emph{Is it sufficient to merely learn an invariant representation and minimize source risks to guarantee a good performance in a new environment? What are the sufficient conditions to guarantee a small test-environment error?} 

\paragraph{Contributions}
In this paper, we aim to address these fundamental problems in domain generalization. Concretely, (1) We reveal the limitation of representation learning in domain generalization through barely ensuring invariance criteria, which can lead to a \emph{over-matching} on the observed environments. e,g the complex or non-smooth representation function will be vulnerable to an unseen distribution-shift; (2) We derive novel theoretical analysis to upper bound the unseen test environment error in the context of representation learning, which highlights the importance of controlling the complexity of the representation function. We further formally demonstrate the Lipschitz property as the sufficient conditions to ensure the smoothness of the representation function; (3) In practice, we propose the \emph{Jacobian matrix regularization} as a new criteria in various invariance criteria and datasets, and the empirical results suggest an improved performance in predicting the test environment.

\section{Background and Motivation}
Throughout this paper, we have $T$ observed (source) environments $\calS_1(x,y) ,\dots,\calS_T(x,y)$ with $x\in\calX$, $y\in\calY$. The goal of domain generalization is to learn a proper representation $\phi:\calX\to\calZ$ and classifier $h:\calZ\to\calY$ to have a good performance on the (unseen) test environment $\calT(x,y)$. 

Specifically, let $\calL$ denote the prediction loss, domain generalization can be formulated as minimizing the following loss:
\begin{equation}
    \min_{\phi,h} \sum_{t} \E_{(x,y)\sim\calS_t} \calL(h \circ \phi(x), y) + \lambda_0~ \text{INV}(\phi,\calS_1,\dots,\calS_T) 
    \label{eq:1}
\end{equation}
where $\text{INV}(\phi,\calS_1,\dots,\calS_T)$ is an auxiliary task to ensure the invariance among the observable source environments, which have various forms:
\begin{enumerate}
    \item \emph{Marginal feature invariance} \cite{ganin2016domain} through enforcing \\
    $\E_{x_1\sim\calS_1(x)}[\phi(x_1)] = \dots = \E_{x_t\sim\calS_t(x)}[\phi(x_t)] = \dots =  \E_{x_T\sim\calS_T(x)}[\phi(x_T)]$, $\forall t\in \{1,\dots,T\}$.  
    \item \emph{Feature conditional invariance} \cite{zhang2013domain} through enforcing $\E_{x_1\sim\calS_1(x|Y=y)}[\phi(x_1)|Y=y] = \dots = \E_{x_t\sim\calS_t(x|Y=y)}[\phi(x_t)|Y=y] = \dots =  \E_{x_T\sim\calS_T(x|Y=y)}[\phi(x_T)|Y=y]$, $\forall t, y$.  
    \item \emph{Label conditional invariance} \cite{arjovsky2019invariant,kamath2021does} through enforcing $\E_{x_1\sim\calS_1(x)}[y|\phi(x_1)=z] = \dots = \E_{x_t\sim\calS_t(x)}[y|\phi(x_t)=z] = \dots =  \E_{x_T\sim\calS_T(x)}[y|\phi(x_T)=z]$, $\forall t, y$.  
\end{enumerate}

The aforementioned invariance principles have been broadly adopted in domain generalization with various empirical algorithms. However, the following counter-examples reveal that merely optimizing  Eq.~(\ref{eq:1}) with different invariance criteria is not a sufficient condition for guaranteeing a reliable prediction in the unseen (test) environment.

\begin{figure}[t]
  \centering
  \begin{subfigure}{0.3\textwidth}
  \centering
     \includegraphics[width=1\textwidth]{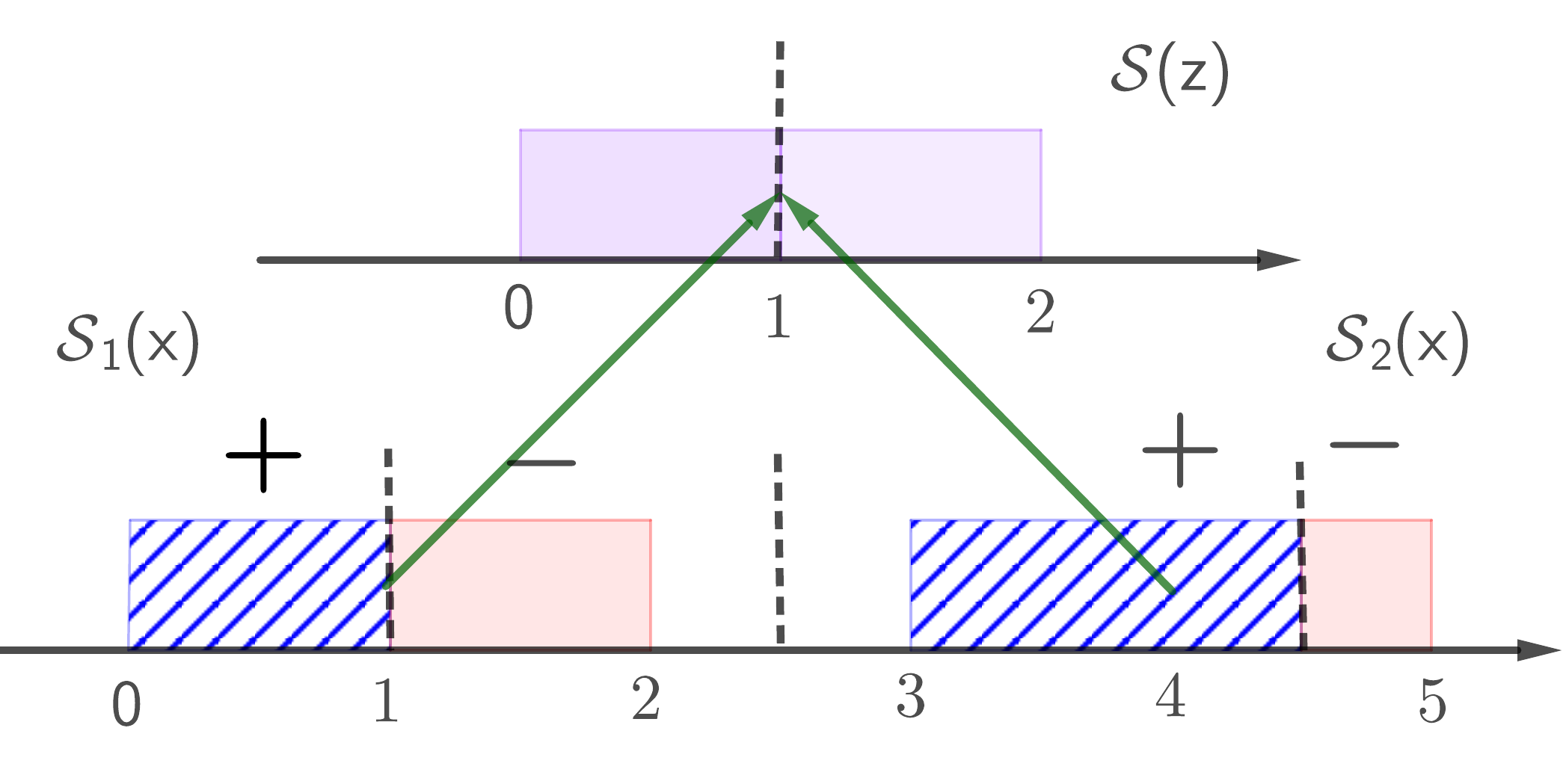}
     \caption{Marginal Invariance}
  \end{subfigure}
  \hfill
 \begin{subfigure}{0.65\textwidth}
  \centering
     \includegraphics[width=0.45\textwidth]{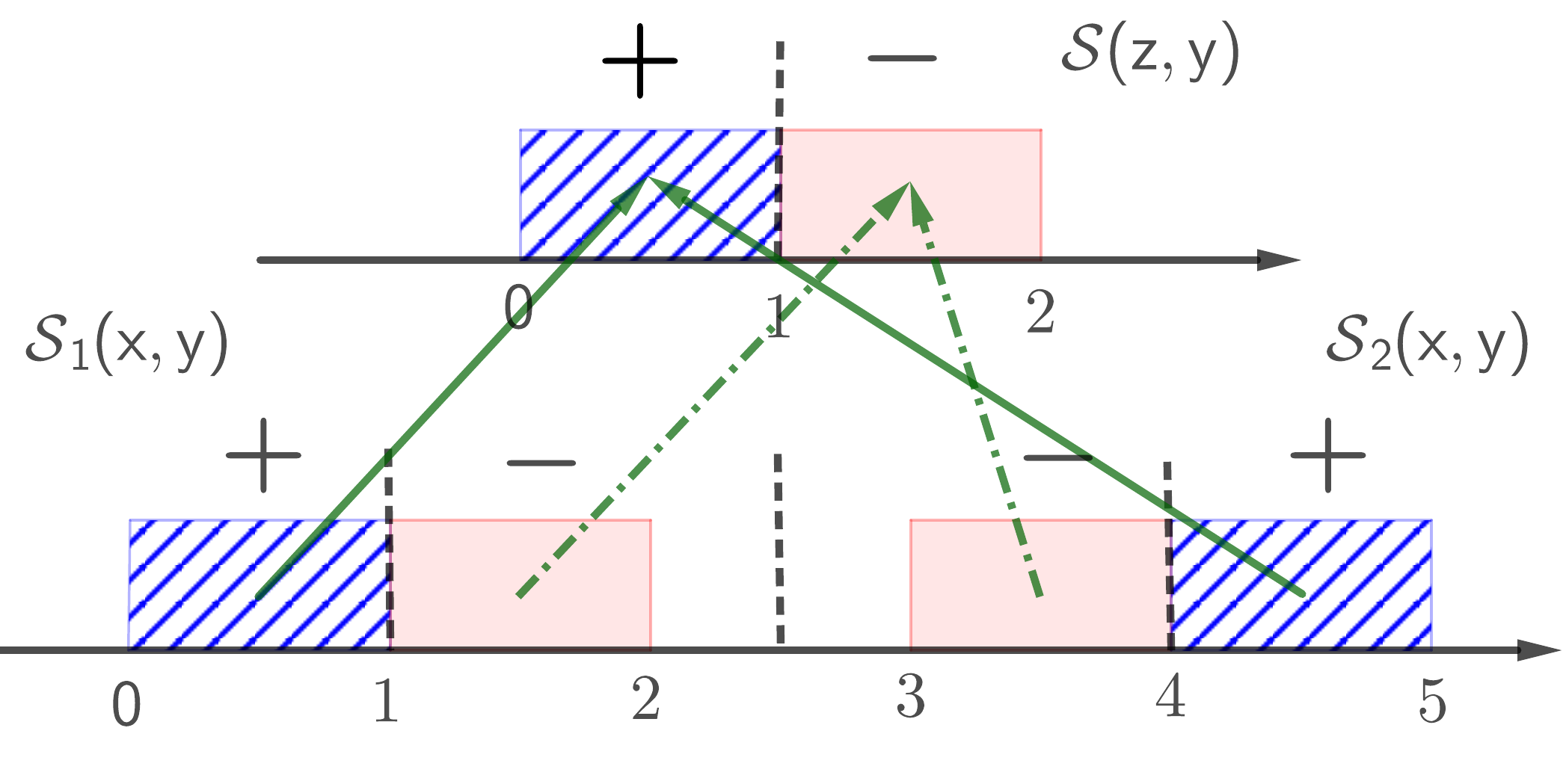}
     \hfill
     \includegraphics[width=0.45\textwidth]{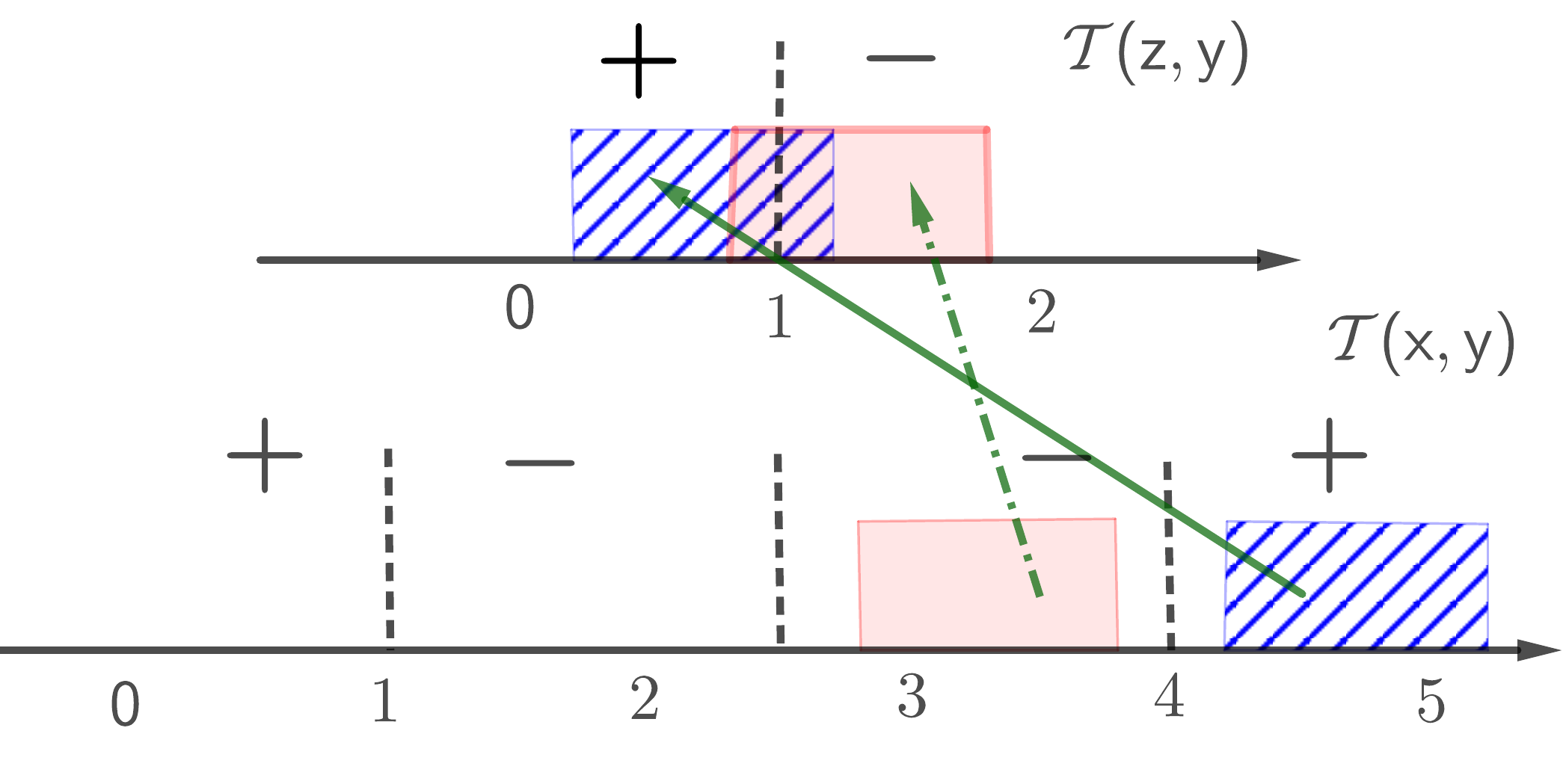}
     \caption{Conditional Invariance: Training (left) and Test (right) environment}
  \end{subfigure}
  \caption{Limitations of optimizing Eq.~(\ref{eq:1}) with different invariance criteria. Intuitively, the marginal invariance fails when conditional shift occurs. The conditional invariance learns an \emph{over-matched} embedding with a non-ignorable prediction error in the test environment.}
  \label{fig:limit_invra}
\end{figure}

In Fig.~\ref{fig:limit_invra}, we illustrated the limitations of these three invariance principles. Specifically: 

(1) Enforcing marginal invariance is problematic when the conditional distributions are different. In Fig.~\ref{fig:limit_invra}(a), two observed environments have different label portions. A simple linear embedding function $\phi$ can ensure $\calS_1(z)=\calS_2(z)$. However, when we adopt a shared classifier $h$, the output prediction distribution $\hat{y}=h(z)$ are identical. Clearly, it is problematic since the label distributions between the environment can be significant different.

(2) Compared with marginal invariance, feature and label conditional invariances impose stronger principles. However, the prediction can be still vulnerable in the test environment due to the \emph{over-matching}. Specifically, in Fig.~\ref{fig:limit_invra}(b, Left), if we adopt the embedding function $\phi$ and classifier $h$ as:
\[\phi(x) =  \begin{cases} 
      x   & 0 \leq x \leq 2 \\
      x-2 & 3\leq x\leq 4 \\
      5-x & 4 < x \leq 5 
  \end{cases}, \quad\quad h(z) =  -\mathrm{sign}(z-1). \]
  
Then, in the latent space $z$, $\forall y \in \calY$ we have the conditional invariance with $\calS_1(y|z)=\calS_2(y|z)$ and $\calS_1(z|y)=\calS_2(z|y)$ and zero prediction error in the observed environments with $\E_{(x,y)\sim\calS_{t}}\calL(h\circ \phi(x),y) = 0$. However, in the test time, if the unseen environment has a consistent shift in Fig.~\ref{fig:limit_invra}(b, Right) such that $\forall y$, $d_{\mathrm{TV}}(\calT(x|Y=y)\|\calS_2(x|Y=y))=\epsilon$ with $0 < \epsilon < 0.5$, then the prediction error w.r.t. (0-1) binary loss is $\E_{(x,y)\sim\calT}\calL(h\circ \phi(x),y) = \epsilon$, which is vulnerable and non-ignorable in the consistent distribution shift. Moreover, this problem can be much more severe in high-dimensional dataset and over-parametrized deep neural network.

The limitation of Eq (1) is the potential \emph{over-matching} in the embedding function, where there exist infinite $\phi$ to minimize Eq.~(\ref{eq:1}) in Fig.~\ref{fig:limit_invra}(b). However, some embedding are rather complex which are poorly generalized to the new environment. In fact, only a subset of $\phi$ are more robust for the consist environment shift, which suggests a proper model selection w.r.t. $\phi$:
\begin{equation}
    \min_{\phi,h} \sum_{t} \E_{(x,y)\sim\calS_t} \calL(h \circ \phi(x), y) + \lambda_0~ \text{INV}(\phi,\calS_1,\dots,\calS_T) + \lambda_1 \textcolor{blue}{\text{Model\_Select}(\phi)}. 
    \label{eq:2_reg}
\end{equation}

In the follow sections, we will derive theoretical results to demonstrate the influence of model selection w.r.t. $\phi$.

\section{Theoretical Analysis}
We aim at proposing a formal understanding of the regularization term in predicting the unseen test environment. Let the embedding being a random transformation (or transition probability kernel) $\Phi(z|x):\calX\to\calZ$, where the deterministic representation function is a special case with $\Phi(z|X=x) = \delta_{\phi(x)}$, where $\delta$ is the delta function. The conditional distribution defined on the latent space $\calZ$ is denoted as $\calS(z) = \int\Phi(z|x)\calS(x)dx$ and $\calS(z|Y=y) = \int\Phi(z|x)\calS(x|Y=y)dx$. Before presenting the theoretical results, there are two additional elements to be clarified:

\paragraph{Performance Metric} Throughout this paper we use \emph{Balanced Error Rate} (BER) rather than the conventional ERM to measure the performance since the training and test environments can be highly label-distribution imbalanced. Specifically, the prediction risk w.r.t. the classifier $h$ and embedding distribution $\Phi$ is
\[
\text{BER}_{\D}(h,\Phi) = \frac{1}{|\calY|}\sum_{y} \E_{z\sim\D(z|Y=y)} \calL(h(z),y)
\]
Intuitively, BER measure the uniform-average classification error on each class.
\paragraph{Invariance Criteria} In our analysis, we mainly focus on the the feature-conditional invariance since the label information is generally discrete or low-dimensional, which is relatively straightforward to realize in practice. We will further justify the feature-conditional invariance can also induce the label-conditional invariance and marginal invariance, shown in Lemma 1.

Based on these two elements, we can demonstrate the risk of test environment in the context of representation learning. 
\begin{theorem}
Supposing \quad 
\begin{enumerate}[label=\roman*),leftmargin=*]
\item observed source environments are $\calS_1(x,y),\dots,\calS_T(x,y)$ and unseen test environment is $\calT(x,y)$;
\item the prediction loss $\calL$ is bounded in $[0,1]$;
\item the embedding distribution $\Phi$ satisfies a small feature-conditional total variation distance on the \textcolor{blue}{latent space $\calZ$}: $\forall i,j \in\{1,\dots,T\}~ y\in\calY$,  $d_{\mathrm{TV}}(\calS_i(z|Y=y)\|\calS_j(z|Y=y))\leq \kappa$;
\item $\forall y \in \calY$, on the \textcolor{blue}{raw feature space $\calX$}: $\underset{t\in\{1,\dots,T\}}{\min}~d_{\mathrm{TV}}(\calT(x|Y=y)\|\calS_t(x|Y=y)) \leq \epsilon$. 
\end{enumerate}
Then the Balanced Error Rate on the test environment is upper bounded by:
\[ 
\text{BER}_{\calT}(h,\Phi) \leq \frac{1}{T} \sum_{t=1}^T \text{BER}_{\calS_t}(h,\Phi) + \kappa + \alpha_{\mathrm{TV}}(\Phi)\epsilon
\]
Where $\alpha_{\mathrm{TV}}(\Phi)$ is Dobrushin coefficient \cite{Polyanskiy2019}: $\alpha_{\mathrm{TV}}(\Phi) :=\sup_{x,x^{\prime}\in\calX} d_{\mathrm{TV}}(\Phi(\cdot|x)\|\Phi(\cdot|x^{\prime}))$
\end{theorem}

\textbf{Discussions} The prediction risk of an unseen test environment is controlled by the following terms:

\begin{wrapfigure}[10]{r}{0.3\textwidth}
\vspace{-0.5cm}
\centering
\includegraphics[width=0.25\textwidth]{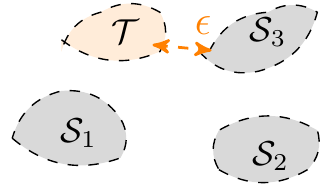}
\caption{Illustration of $\epsilon$: distance between $\calT$ and its nearest source $\calS_3$.}
\label{fig:fig2}
\end{wrapfigure}
(1)~The first term suggests to learn $h$ and $\Phi$ to minimize the BER over the labeled data from the source environments;

(2)~A small $\kappa$ indicates learning $\Phi$ to match feature-conditional distribution. Specifically, when $\kappa=0$, we have $\calS_1(z|Y=y)=\dots=\calS_T(z|Y=y)$, achieving  feature-conditional invariance;

(3)~$\epsilon$ in the third term is a \emph{unobservable} factor in the learning. As Fig.~\ref{fig:fig2} shows, $\epsilon$ reveals the inherent relations between the test and source environments. Intuitively, a small $\epsilon$ indicates the test environment $\calT$ is similar to one of the observed sources, which indicates we can more easily predict the test through leveraging the knowledge from the sources;

(4)~$\alpha_{\mathrm{TV}}(\Phi)$ in the third term is the \emph{controllable} factor as a regularization of $\Phi$. Specifically, $\alpha_{\mathrm{TV}}(\Phi)$ reflects the the smoothness of the embedding distribution. At the test time, regularization on $\Phi$ is crucial since the $\epsilon$ is \emph{unknown, uncontrollable and even non-ignorable}. That is, merely minimizing Eq.~(\ref{eq:1}) by ensuring $\text{BER}_{\calS_t}(h,\Phi)=0$ and $\kappa=0$ are not sufficient. If $\alpha_{\mathrm{TV}}(\Phi)$ is large, the upper bound will become vacuous and generalization in the test environment is not necessarily guaranteed;

(5)~The trade-off in learning $\Phi$. Although $\alpha_{\mathrm{TV}}(\Phi)$ suggests a smooth representation, however over-smoothing is harmful in learning meaningful representation. For instance, if embedding distribution $\Phi$ is a constant, then $\alpha_{\mathrm{TV}}(\Phi)=0$, the network does not learn an embedding and $\text{BER}_{\calS_t}(h,\Phi)$ will be inherently large. 
\\ \\
Compared with most previous theoretical results, our results highlight the role of representation learning in domain generalization. In particular, Theorem 1 further motivates novel algorithm to control the Dobrushin Coefficient, which is shown in Sec 3.2 and Sec 4.   

\subsection{Relation with other invariance criteria}
Theorem 1 justifies the importance of considering regularizing of $\Phi$ under feature-conditional invariance, the following Lemma reveals the relations with other two invariance criteria.

\begin{lemma}
If the embedding distribution $\Phi$ satisfies a small feature-conditional total variation distance on the \textcolor{blue}{latent space $\calZ$}: $\forall i,j \in\{1,\dots,T\}~ y\in\calY$, $d_{\mathrm{TV}}(\calS_i(z|Y=y)\|\calS_j(z|Y=y))\leq \kappa$ and $\calS_i(Y=y)=\calS_j(Y=y)=\frac{1}{|\calY|}$, then we have
\[
\E_{z\sim\Omega^{\star}} |\calS_i(y|z)-\calS_j(y|z)|\leq C^{+}\kappa, \quad \E_{z\sim\Omega^{\star}} |\calS_i(z)-\calS_j(z)|\leq \kappa,
\]
where $C^{+}$ is a positive constant and $\Omega^{\star} = \text{supp}(\calS_i(z))\cap\text{supp}(\calS_j(z))$ denotes the intersection of latent space between two environments.
\end{lemma}
Lemma 1 reveals that the feature-conditional invariance can induce other two types of invariances if the label distribution among the source is balanced, which is practically feasible through re-sampling the dataset as uniform distribution. Specifically, if $\kappa=0$, we can achieve other two invariances.

\subsection{Sufficient conditions for controlling Dobrushin Coefficient}
We will discuss sufficient conditions for controlling the Dobrushin Coefficient, which is intuitively interpreted as smoothness properties of representation. Lemma 2 shows that a Lipschitz condition is \emph{one} sufficient condition to control $\alpha_{\mathrm{TV}}(\Phi)$.  

\begin{lemma}
Supposing the embedding distribution $\Phi(z|x)$ is $d$-dimensional parametric Gaussian distribution with $z\sim \calN (\phi(x),\sigma^2\mathbf{I}_{d})$ and $d_{\max} = \sup_{x,x^{\prime}\in\calX}~\|x-x^{\prime}\|_2$, then the Dobrushin Coefficient can be upper-bounded by:
\[
\alpha_{\mathrm{TV}}(\Phi) \leq \sqrt{2}\left(1- \exp(-\frac{d^2_{\max}}{8d \sigma^2} L^2_{\phi} )\right)^{1/2} 
\]
where $L_{\phi}$ is the Lipschitz constant of $\mu_{\phi}(x)$, i.e  $\forall x,x^{\prime}\in\calX, \|\phi(x)-\phi(x^{\prime})\| \leq L_{\phi} \|x-x^{\prime} \|_2$. 
\end{lemma}
We can verify that if $L_{\phi}\to 0$, then $\alpha_{\mathrm{TV}}(\Phi) \to 0$. In the conventional deep neural-network, the deterministic parametric embedding can be approximated as the mean ($\phi(x)$) of the conditional distribution with a small variance \cite{achille2018emergence}.
Therefore, Lemma 2 suggests learning a Lipschitz embedding to promote a better generalization property in the test environment $\calT$.

\section{Practical Implementations}
We have demonstrated the Lipschitz property of embedding function $\phi$ can induce a better generalization property. In this section, we will further elaborate practical implementations to realize the Lipschitz property of the embedding function through multiple observed source environments. 

\begin{wrapfigure}[7]{r}{0.28\textwidth}
\centering
\includegraphics[width=0.25\textwidth]{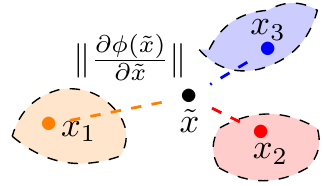}
\caption{Illustration of the virtual sample regularization.}
\label{fig:fig3}
\end{wrapfigure}

It has been proved that the Frobenius norm of Jacobian matrix w.r.t $\phi$ is the upper bound of small Lipschitz constant of $\phi$ \cite{miyato2018spectral}. 
In order to take advantage of multiple-environments, we create virtual samples $\tilde{x}$ through a linear combination of the samples from the sources, shown in Fig.~\ref{fig:fig3}. The linear combination coefficients $(\gamma_1,\dots,\gamma_T)$ are generated through the Dirichlet distribution with hyper-parameter $\beta=1$. The aim of creating virtual samples $\tilde{x}$ is to \emph{enforce a smooth prediction behavior on the unobserved regions between the environments}, which can be broadly viewed as data-augmentation based approach. (This will be discussed in the related work.)

\begin{algorithm}[h]
\caption{Regularization of $\phi$}
\label{algo1}
\begin{algorithmic}[1]
\Require Multiple-source data-sets $\calS_1,\dots,\calS_T$, embedding $\phi$, hyper-parameter $\beta$.
\State $x_1\sim\calS_1(x),\dots,x_T\sim\calS_T(x)$, $(\gamma_1,\dots,\gamma_T)\sim\text{Dirichlet}(\beta,\dots,\beta)$ \Comment{Sampling}
\State $\tilde{x} = \sum_{t=1}^{T} \gamma_t x_t $ \Comment{Create virtual samples}
\State \textbf{return} $\|\frac{\partial \phi(\tilde{x})}{\partial \tilde{x}}\|_{F}$ \Comment{Compute Frobenius Norm of Jacobin matrix}
\end{algorithmic}
\end{algorithm}

\textbf{Regularization is independent of learning invariance}
We denote the \emph{black-box} algorithms that achieve the invariance (e.g. feature, label and feature conditional invariance) as $\text{INV}(\phi,\calS_1,\dots,\calS_T)$, which includes a board range of algorithms. Then the improved loss can be expressed as:
\[ \min_{\phi,h} \frac{1}{T}\sum_{t}\mathrm{BER}_{\calS_t}(h\circ\phi) + \lambda_0~ \text{INV}(\phi,\calS_1,\dots,\calS_T) + \lambda_1 \E_{\tilde{x}} \|\frac{\partial \phi(\tilde{x})}{\partial \tilde{x}}\|_{F}. 
\]
In the experimental part, we will investigate different invariance approaches and the benefits of the regularization.

\section{Related Work}

\paragraph{Learning invariance} is a popular and widely adopted approach in domain generalization. Inspired from the techniques in deep domain adaptation \cite{ben2010theory}, various approaches have been proposed to enable different invariance criteria such as marginal invariance $\calS_1(z)=\dots=\calS_T(z)$ \cite{ganin2016domain,li2018domain,sicilia2021domain,albuquerque2019generalizing}.  However, the proposed theoretical results are mainly inspired from unsupervised domain adaptation, which does not consider the specific scenarios in domain generalization. i.e, the label information is known during the alignment, which can induce better alignments.  As for feature conditional invariance $\calS_1(z|y)=\dots=\calS_T(z|y)$ \cite{li2018learning,wang2020domainmix,zhao2020domain,ilse2019diva}, it considers the label information and enforce stronger conditions among the sources. However, as our counterexample indicates, merely learning the conditional invariance is not sufficient to provably guarantee the unseen test prediction risk. In contrast, we further formally reveal the limitation of representation learning w.r.t. conditional invariance, which remains elusive in the previous work.
A more recent approach is to learn label conditional invariance, i.e. ensuring the same decision boundary across the different environments (IRM \cite{arjovsky2019invariant,lu2021nonlinear}). However, recent work reveals the failure scenarios in IRM, which can be explained through our theoretical analysis.

\paragraph{Relation with Data-Augmentation Based Approach}
It has been recently observed that data-augmentation based approaches are quite effective in various practical domain generalization \cite{volpi2018generalizing,li2019feature,zhou2020learning,zhou2021domain,muller2020learning}. Intuitively, augmentation based approaches aim at generate new samples from observed environments to enable smoother prediction results. In this part, we aim to prove the role of data-augmentation, which is \emph{implicit} to learn a smooth representation and consistent with our theoretical results.

Specifically, we consider one typical case with a conditional \emph{black-box} interpolation function $\text{INP}$ with $\tilde{x} = \text{INP}(x_1,\dots,x_T; y)$ with $x_1\sim\calS_1(x|y), \dots, x_T\sim\calS_T(x|y)$. For instance, considering object classification under different background, the conditional augmentation aims at creating the same object through considering information from different environments. We further suppose the binary classification problem with $\calY=\{-1,+1\}$, the classifier is linear with $h(z)=w^T z$ and the prediction loss is logistic loss with $\calL(\hat{y},y)= \log(1+\exp(-\hat{y}y))$. The the augmentation loss can be written as:
\[
R_{\text{aug}} = \sum_{y} \E_{\tilde{x}\sim\text{INP}(x_1,\dots,x_T; y)} \calL(w^T\phi(\tilde{x}), y)
\]
If we use second-order Taylor approximation at $\E_{\tilde{x}} [\phi(\tilde{x})]$, the centroid of the augmentation feature on the embedding space, then the prediction loss can be approximated as:
\[
R_{\text{aug}} \approx \sum_{y} \underbrace{\calL(w^{T} \E_{\tilde{x}}[\phi(x)], y)}_{(1)} + \underbrace{\frac{1}{2} \E_{\tilde{x}} [(w^{T} (\phi(\tilde{x})-\E_{\tilde{x}} [\phi(\tilde{x})]))^2 \calL^{\prime\prime}(w^{T}\E_{\tilde{x}} [\phi(\tilde{x})],y)}_{(2)}
\]
The augmented prediction loss can be approximated two terms: (1) suggests a small loss on the centroid of the generated feature, (2) indicates a smooth prediction on the new generated sample. Since $\calL^{\prime\prime}(w^{T}\E_{\tilde{x}} [\phi(\tilde{x})],y)\leq 1$ and $\phi$ is Lipschitz function, (2) can be further upper-bounded by:
\[
\text{(2)} \leq L^2_{\phi} \frac{\|w\|^2_2}{4}\text{Var}(\tilde{x})
\]
Therefore, if an embedding function is set to be smooth with a small lipschitz constant, the second order approximation of the augmentation loss can be controlled. Therefore, minimizing the prediction loss on the augmented data can be viewed as an implicit approach to enable the smooth representation.

\section{Experiments}
In the experimental part, we aim to address the following question:

\emph{Is the regularization term effective to generalize in the unseen environments? In what are scenarios that the regularization is beneficial ?}

\subsection{Choice of Invariance Criteria and Loss}
We evaluate the proposed regularization through typical invariance representation algorithms to verify the effectiveness of the regularization. \\
(1) \textbf{DANN}\cite{ganin2016domain} aims at enforcing the invariance w.r.t. $\calS_1(z)=\dots=\calS_T(z)$ through min-max optimization. Concretely, we introduce a domain discriminator $d:\calZ\to\{1,\dots,T\}$, such that 
\[ 
\min_{\phi} \text{INV}(\phi,\calS_1,\dots,\calS_T) = \min_{\phi}\max_{d}~\frac{1}{T} \sum^{T}_{t=1}  \E_{x_t \sim \calS_t(x)} \mathbf{1}_{t} \log(d\circ\phi(x_t))
\]
Where $\mathbf{1}_{t}$ is the one-hot vector. Intuitively, the discriminator tried to minimize the cross-entropy loss to differentiate the sources and ensure the embedding to learn an invariant representation. \\
(2) \textbf{Feature-Conditional Invariance (CDANN)} Adapted from \cite{mirza2014conditional,li2018domain}, we aim at enforcing  
$\calS_1(z|y)= \dots =\calS_T(z|y)$. We introduce a conditional domain discriminator $d:\calZ\times\calY\to\{1,\dots,T\}$, such that:
\[ 
\min_{\phi} \text{INV}(\phi,\calS_1,\dots,\calS_T) = \min_{\phi}\max_{d}~\frac{1}{T} \sum^{T}_{t=1}  \E_{(x_t,y_t) \sim \calS_t(x,y)} \mathbf{1}_{t} \log\left(d\circ(\phi(x_t)\otimes y_t)\right)
\]
(3) \textbf{Label-Conditional Invariance (IRM)} \cite{arjovsky2019invariant} proposed a regularization term to encourage the $\calS_1(y|z)= \dots =\calS_T(y|z)$. Specifically, they assume the predictor equals to $1$ with 
\[ 
\min_{\phi} \text{INV}(\phi,\calS_1,\dots,\calS_T) = \min_{\phi}~\frac{1}{T} \sum^{T}_{t=1} \|\nabla_{h|h=1} \E_{\calS_t} \calL(h\circ\phi(x_t),y_t)\|^2 
\]

As for $\calL$, we adopted the cross-entropy as the prediction loss.

\subsection{Dataset description and Experimental setup}
The experiment validation consists in evaluating toy and real-world datasets to verify the effectiveness of the regularization.

\textbf{ColorMNIST} \cite{arjovsky2019invariant} Each MNIST image is either colored by red or green, in order to strongly correlate (but spuriously) with the class label. Thus the class label is strongly correlated with the color than with the digit configuration. The algorithm purely minimizing training error will tend to exploit the false relation of the color, which will lead to a poor generalization in the unseen distribution with different color relations. 
    
Following \cite{arjovsky2019invariant}, the dataset is constructed as follows.
(1) \emph{Preliminary binary label}.  We randomly select 5K samples from MNIST and construct preliminary binary label $\tilde{y}=0$ for digits 0-4 and $\tilde{y}=1$ for 5-9; (2) \emph{Adding label noise}.  We obtain the final label $y$ by flipping $\tilde{y}$ with probability 0.25; (3) \emph{Adding color as spurious feature}. We add the color to the gray-scale digit image by flipping $y$ with probability $P_{\calS}$ (i.e, coloring $y=1$ with red and $y=0$ with green by probability $1-P_{\calS}$).

The ColorMNIST creates a \emph{controllable} environment through assigning various $P_{\calS}$, which enable us to evaluate the generalization performances under different unobserved environments. 

\textbf{PACS} \cite{li2017deeper} and \textbf{Office-Home} \cite{venkateswara2017deep} are real-world datasets with high-dimensional images.  In PACS, the dataset consists four domains Photo (P), Art (A), Cartoon (C), Sketch (S) with 7  classes. In Office-Home, the dataset includes four domains Art (A), Clipart (C), Product (P) and Real World (R) with 65 classes. 

\paragraph{Experimental Setup} We use the standard domain generalization framework DomainBed \cite{gulrajani2021in} to implement our algorithm. In ColorMNIST, we adopt the LeNet structure with three CNN layers as $\phi$ and three fc-layers as $h$. The mini-batch is set as 128 with Adam optimizer with $\lambda_0 = 1$, $\lambda_1\in[10^{-3},1]$. In PACS and Office-Home datasets, we adopt the pre-trained ResNet-18 as $\phi$ and three fc-layers as $h$. We adopted training-domain validation set \cite{gulrajani2021in} to search the best hyper-parameter configuration. Specifically, we 
set the batch size as 64 and $\lambda_0 \in [10^{-7},10^{-2}]$ and $\lambda_1 \in [10^{-5},1]$.  We adopt the train-validation split approach (i.e, we randomly split the observed environment as training and validation set and tune the best configuration on the validation set w.r.t. the $\calS$. We did not know the test environment during the tuning.) to search the best hyper-parameter. We run the experiments five times and report the average and std.  The detailed network structures are delegated in the appendix.
 
\subsection{Empirical Results}
\begin{table}[h]
  \caption{Empirical Results (Accuracy Per-Class on $\%$, bold indicates a statistical significant result) on ColorMNIST. We have three environments with different $P_{\calS} = \{0.1, 0.2, 0.9\}$. In the domain-generalization, we train on two environments and test on the untrained environment.}
  \label{results:color_mnist}
  \centering
  \begin{tabular}{l|c|c|c|c}
    \toprule
    Method/Test Env   & $P_{\calS} = 0.1$  &  $P_{\calS} = 0.2$ &  $P_{\calS} = 0.9$ & Average \\
    \midrule
    ERM  & 60.2 $\pm$ 0.9 & 65.7 $\pm$ 0.6 & 26.8 $\pm$ 1.8  & 50.9 \\
    ERM+REG & \textbf{65.0} $\pm$ \textbf{1.9} & \textbf{69.4} $\pm$ \textbf{1.6} & \textbf{29.1} $\pm$ \textbf{1.3} & 54.5 \\
    \midrule
    DANN  & 60.3 $\pm$ 2.3 & 66.2 $\pm$ 0.5 & 26.7 $\pm$ 2.5  & 51.1 \\
    DANN+REG & \textbf{68.2} $\pm$ \textbf{1.3} & \textbf{70.9} $\pm$ \textbf{1.7} & 27.9 $\pm$ 2.1 & 55.7 \\
    \midrule
    CDANN      & 62.7 $\pm$ 1.9 & 66.7 $\pm$ 2.0 & 27.1 $\pm$ 3.2  & 52.2 \\
    CDANN+REG  & \textbf{70.3} $\pm$ \textbf{0.5} & \textbf{72.2} $\pm$ \textbf{1.2} & \textbf{30.6} $\pm$ \textbf{1.7} &  57.7 \\
    \midrule
    IRM   & 57.2 $\pm$ 1.7 & 63.3 $\pm$ 2.1 & 40.7 $\pm$ 10.5 & 53.7 \\
    IRM +REG & \textbf{61.9} $\pm$ \textbf{1.6} & \textbf{66.5} $\pm$ \textbf{3.3} & \textbf{51.2} $\pm$ \textbf{1.5}  &  59.9 \\
    \bottomrule
  \end{tabular}
\end{table}

\begin{table}[h]
  \caption{Empirical Results (Accuracy Per-Class on $\%$, bold indicates a statistical significant result) on PACS. We have four environments Photo (P), Art (A), Cartoon (C) and Sketch (S). In the domain-generalization, we train the model on three environments and test on the untrained environment.}
  \label{results:pacs}
  \centering
  \begin{tabular}{l|c|c|c|c|c}
    \toprule
    Method/Test Env   & Art &  Cartoon  &  Sketch  &  Photo & Average \\
    \midrule
    ERM  & 74.2 $\pm$ 1.2 & 71.8 $\pm$ 1.1 & 93.4 $\pm$ 0.9  &  71.4 $\pm$ 0.6 & 77.7\\
    ERM+REG & \textbf{77.4} $\pm$ \textbf{1.4} & 73.1 $\pm$ 0.7 & \textbf{94.8} $\pm$ \textbf{0.8} & \textbf{73.5} $\pm$ \textbf{1.7} &  79.7 \\
    \midrule
    DANN  & 77.3 $\pm$ 1.7 & 74.4 $\pm$ 1.5 & 93.3 $\pm$ 1.1  & 71.7 $\pm$ 2.5 & 79.2\\
    DANN+REG & \textbf{81.1} $\pm$ \textbf{1.6} & \textbf{75.4} $\pm$ \textbf{0.7} & \textbf{94.8} $\pm$ \textbf{1.2} & \textbf{75.8} $\pm$ \textbf{1.1} & 81.6 \\
    \midrule
    CDANN      & 79.6 $\pm$ 2.1 & 75.4 $\pm$ 1.8 & 93.8 $\pm$ 1.2  &  72.3 $\pm$ 1.1 & 80.3 \\
    CDANN+REG  & \textbf{82.5} $\pm$ \textbf{0.5} & \textbf{78.1} $\pm$ \textbf{0.5} & \textbf{95.4} $\pm$ \textbf{0.8} & \textbf{77.0} $\pm$ \textbf{0.8} & 83.3 \\
    \midrule
    IRM   & 69.0 $\pm$ 1.3 & 68.3 $\pm$ 1.7 & 88.7 $\pm$ 2.5 &  64.3$\pm$ 1.2 & 72.6\\
    IRM+REG & \textbf{73.7} $\pm$ \textbf{1.9} & \textbf{70.9} $\pm$ \textbf{2.5} & \textbf{92.1} $\pm$ \textbf{1.3}  & \textbf{67.2} $\pm$ \textbf{2.0} & 76.0 \\
    \bottomrule
  \end{tabular}
\end{table}

\begin{table}[h]
  \caption{Empirical Results (Accuracy Per-Class on $\%$, bold suggests a statistical significant result) on Office-Home. We have four environments Art (A), Clipart (C), Product (P) and Real-world (R). In the domain-generalization, we train the model on three environments and test on the untrained environment.}
  \label{results:office_home}
  \centering
  \begin{tabular}{l|c|c|c|c|c}
    \toprule
    Method/Test Env   & Art &  Clipart  &  Product  &  Real-World  & Average \\
    \midrule
    ERM  & 46.8 $\pm$ 0.9 & 41.2 $\pm$ 0.8 & 64.5 $\pm$ 1.1  &  66.1 $\pm$ 0.7 & 54.7 \\
    ERM+REG & \textbf{48.7} $\pm$ \textbf{0.9} & 42.1 $\pm$ 1.0 & 65.5 $\pm$ 0.7 & 67.1 $\pm$ 0.6 & 55.9\\
    \midrule
    DANN  & 48.0 $\pm$ 0.8 & 44.4 $\pm$ 0.9 & 65.7 $\pm$ 1.2  & 66.5 $\pm$ 0.8 & 56.1 \\
    DANN+REG & \textbf{50.5} $\pm$ \textbf{1.1} & \textbf{46.0} $\pm$ \textbf{0.8} & \textbf{68.0} $\pm$ \textbf{0.8} & \textbf{68.5} $\pm$ \textbf{0.9} &  58.3 \\
    \midrule
    CDANN      & 48.6 $\pm$ 1.1 & 44.7 $\pm$ 0.7 & 65.6 $\pm$ 1.1  &  66.3 $\pm$ 0.8 & 56.3 \\
    CDANN+REG  & \textbf{52.0} $\pm$ \textbf{1.3} & \textbf{47.2} $\pm$ \textbf{0.7} & \textbf{67.9} $\pm$ \textbf{0.8} & \textbf{69.4} $\pm$ \textbf{1.0} & 59.1 \\
    \midrule
    IRM   & 47.2 $\pm$ 0.7 & 42.3 $\pm$ 1.9 & 63.4 $\pm$ 1.5 &  65.3$\pm$ 2.2 & 54.6\\
    IRM+REG & \textbf{49.1} $\pm$ \textbf{1.2} & 43.8 $\pm$ 1.3 & \textbf{66.1} $\pm$ \textbf{1.2}  & \textbf{68.4} $\pm$ \textbf{1.8} & 56.9 \\
    \bottomrule
  \end{tabular}
\end{table}

The results presented in Tab.\ref{results:color_mnist}, \ref{results:pacs} and \ref{results:office_home}. In all datasets and different invariance criteria, the regularization suggest a consistent improvement (ranging from $1.2\%-6.2\%$). Specifically, the improvement in synthetic dataset ColorMNIST is significant, which reveals the effectiveness of the proposed regularization. Moreover, in the real-world datasets such as Office-Home and PACS, the regularization suggests consistent better performance. 

\subsection{Analysis}
We further conduct various analysis to understand the property and role of regularization. 
\paragraph{Influence of regularization}
\begin{figure}[t]
  \centering
  \begin{subfigure}{0.3\textwidth}
  \centering
     \includegraphics[width=0.7\textwidth]{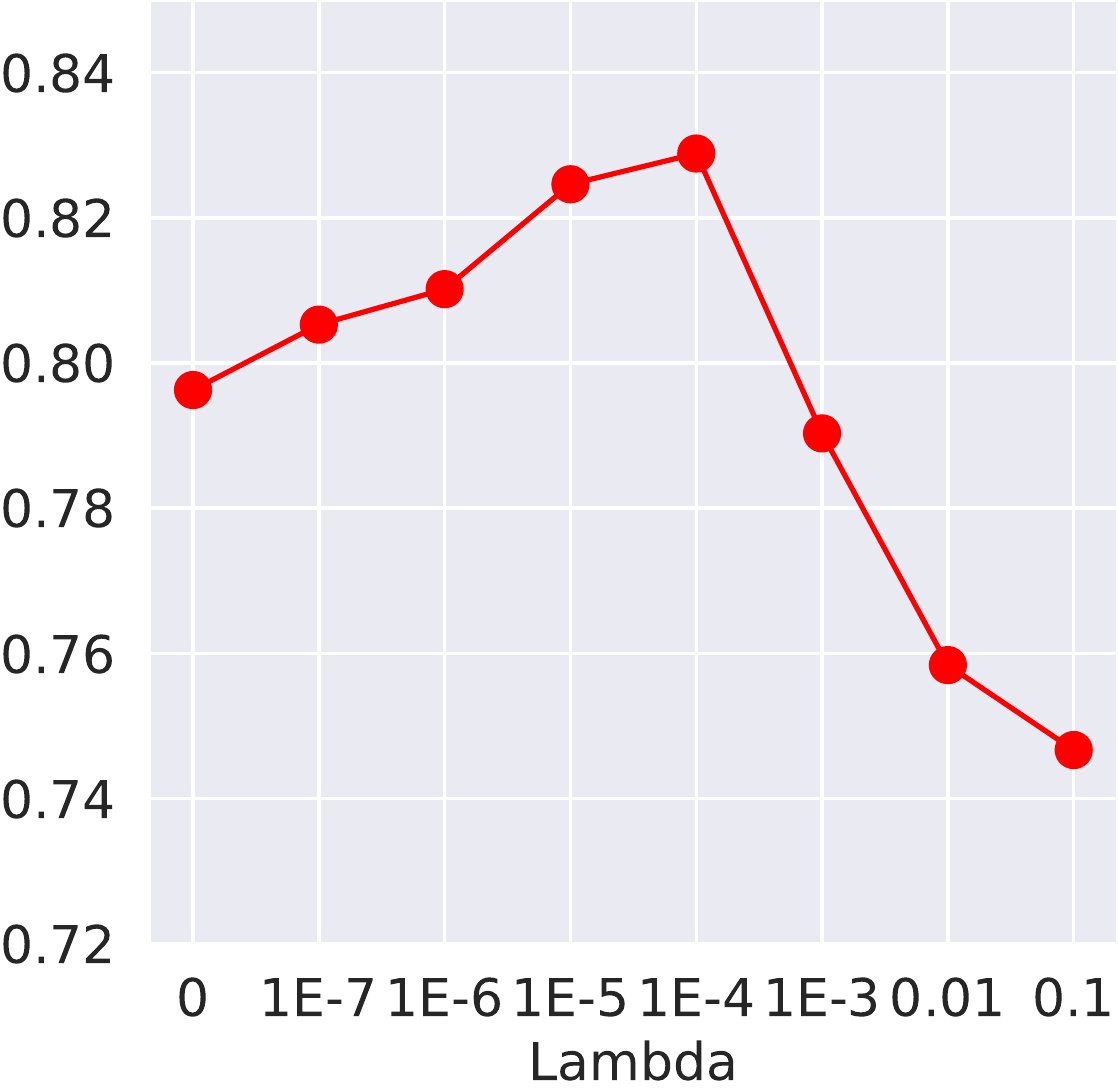}
     \caption{Art}
  \end{subfigure}
  \hfill
 \begin{subfigure}{0.3\textwidth}
  \centering
     \includegraphics[width=0.7\textwidth]{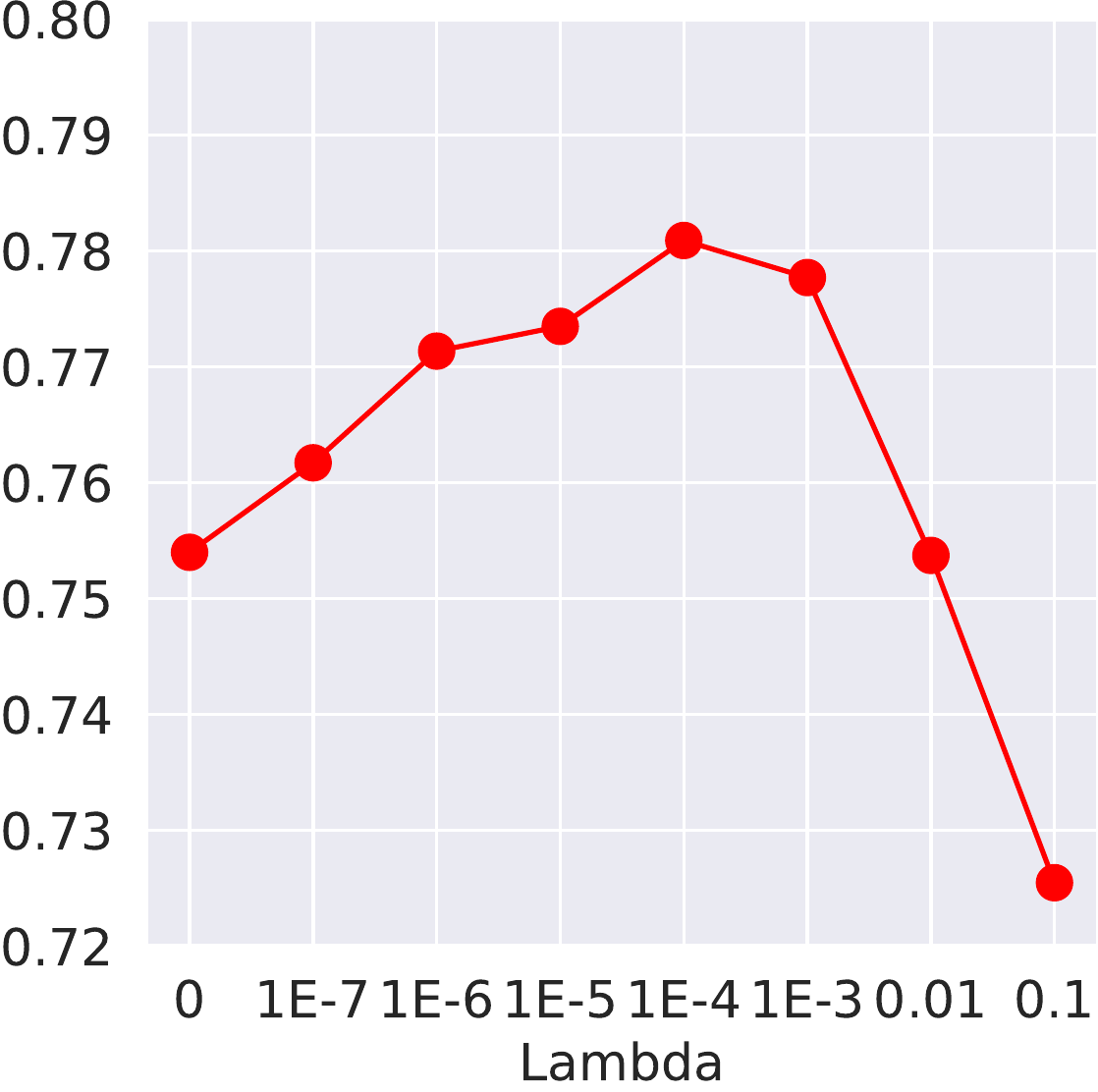}
     \caption{Cartoon}
  \end{subfigure}
  \hfill
  \begin{subfigure}{0.3\textwidth}
  \centering
     \includegraphics[width=0.7\textwidth]{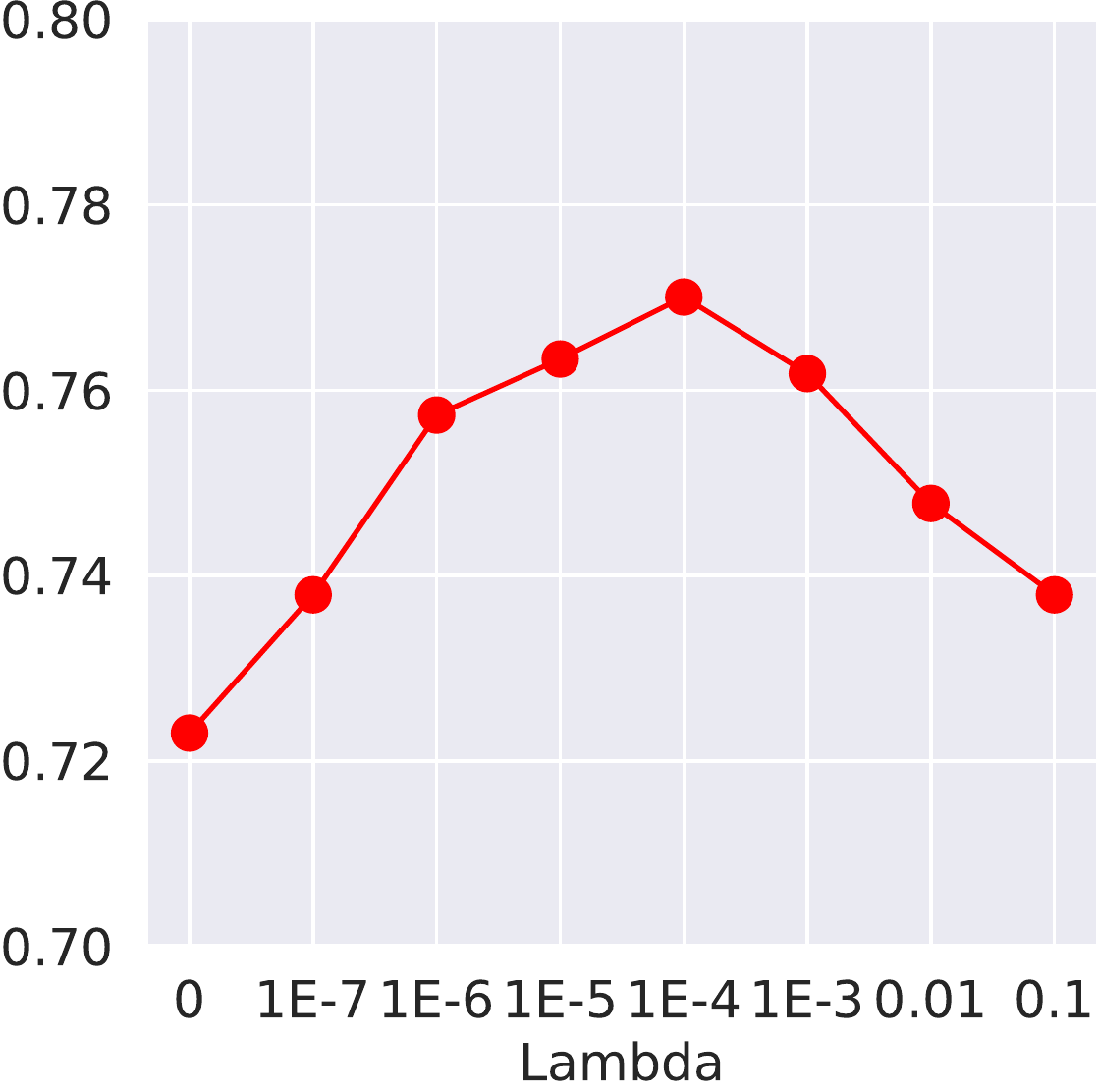}
     \caption{Photo}
  \end{subfigure}
  \caption{Influence of regularization in PACS dataset in CDANN. We gradually change the importance of regularization (i.e, different $\lambda_1$). The accuracy first increases with a larger $\lambda_0$, then the accuracy drops due to a strong smoothing on the representation.}
  \label{fig:abl_study}
\end{figure}
For a better understanding of the regularization, we gradually change $\lambda_1$ to show the influence of regularization. The empirical results are consistent with our theoretical analysis: in the presence of small regularization, the prediction performance can be improved. However, a strong regularization (over smoothing) on the representation learning can be harmful with a dropped prediction performance.

\paragraph{Evolution of Training}
\begin{figure}[t]
  \centering
  \begin{subfigure}{0.4\textwidth}
  \centering
     \includegraphics[width=1.0\textwidth]{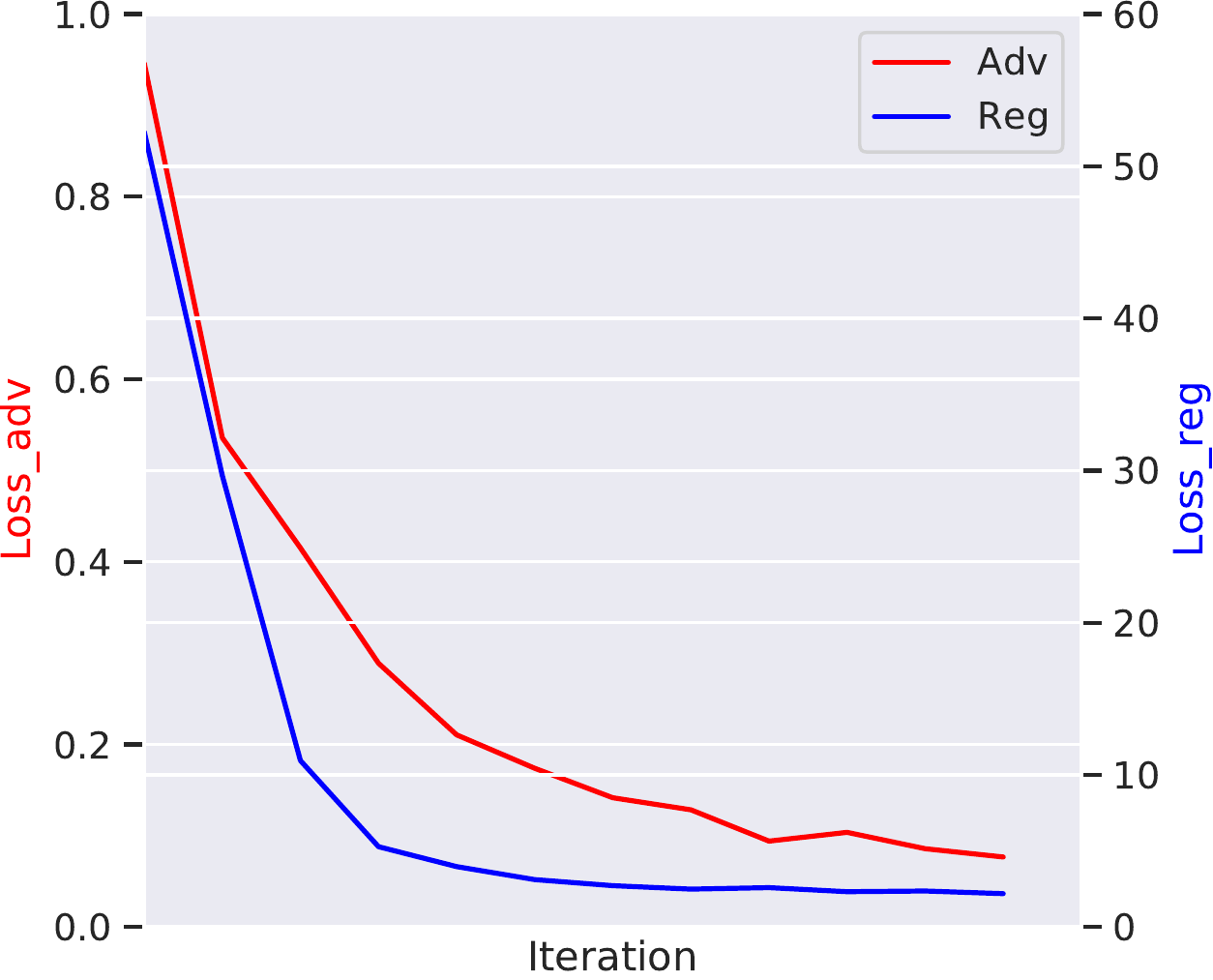}
     \caption{With Regularization}
  \end{subfigure}
  \hfill
 \begin{subfigure}{0.4\textwidth}
  \centering
     \includegraphics[width=1.0\textwidth]{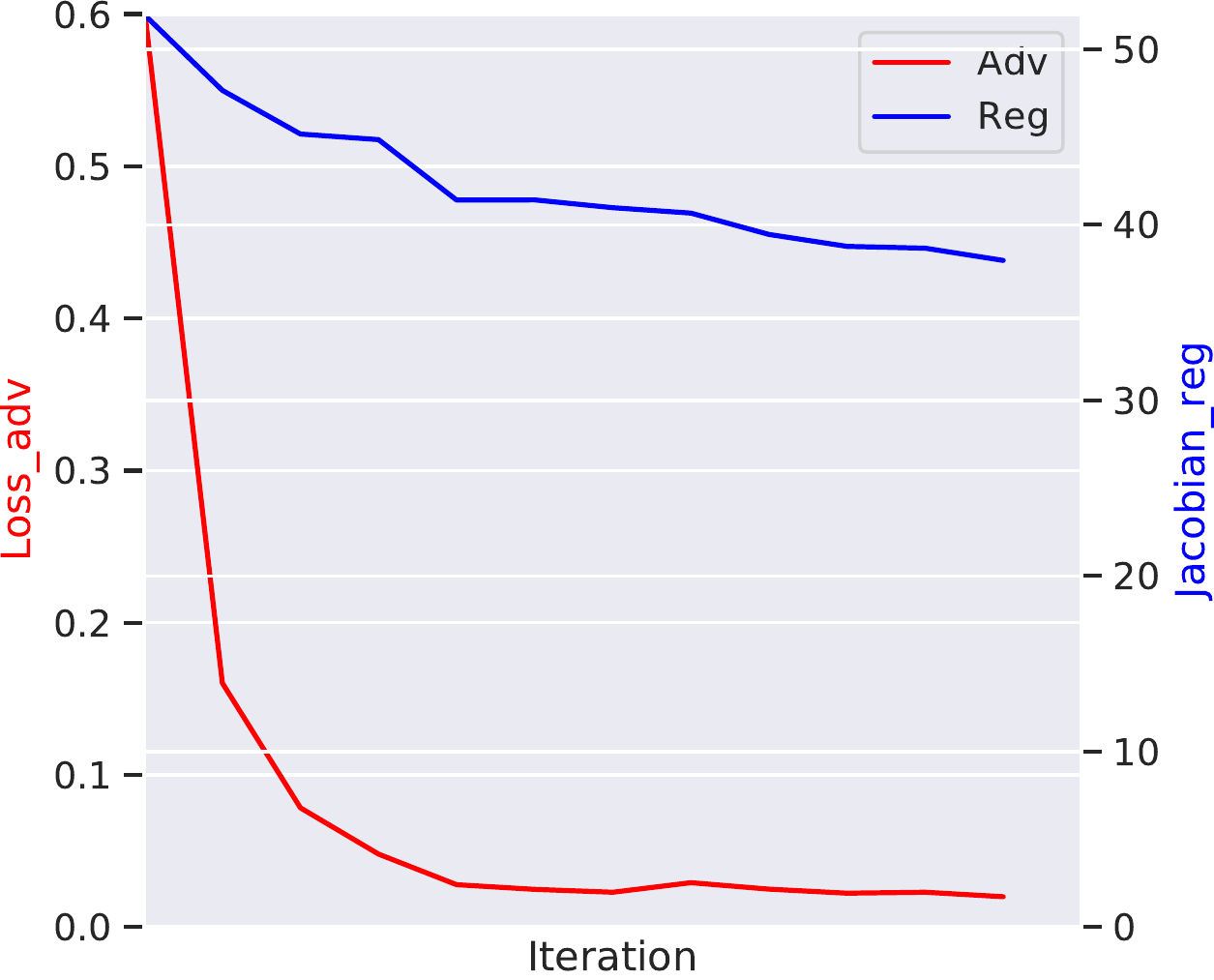}
     \caption{Without Regularization}
  \end{subfigure}
  \caption{Loss Evolution in Office-Home dataset (Training Environments: Clipart, Product, Real-World) in CDANN. \emph{Left}: The evolution of adversarial loss and regularization term if we adopt the regularization loss. \emph{Right}: The evolution of adversarial loss and regularization term (Norm of Jacobin matrix) \textbf{without} adopting regularization loss. The results reveal that without explicit regularization loss, the norm of Jacobin matrix still gradually (but slowly) diminishes. In contrast, adding an explicit term can accelerate the optimization procedure.}
  \label{fig:evo_loss}
\end{figure}
We additionally visualize the evolution of adversarial loss and the norm of Jacobin matrix in two training modes: conditional alignment with and without regularization. Clearly, training without explicit regularization will lead to a relative large norm of Jacobian matrix. During the optimization procedure, the norm of Jacobian matrix gradually but slowly diminishes, which is possibly caused by the implicit regularization through SGD based approach \cite{roberts2021sgd}.  Therefore, adding an explicit regularization term can help a better generalization property.

\begin{figure}[t]
    \centering
    \includegraphics[width=0.5\textwidth]{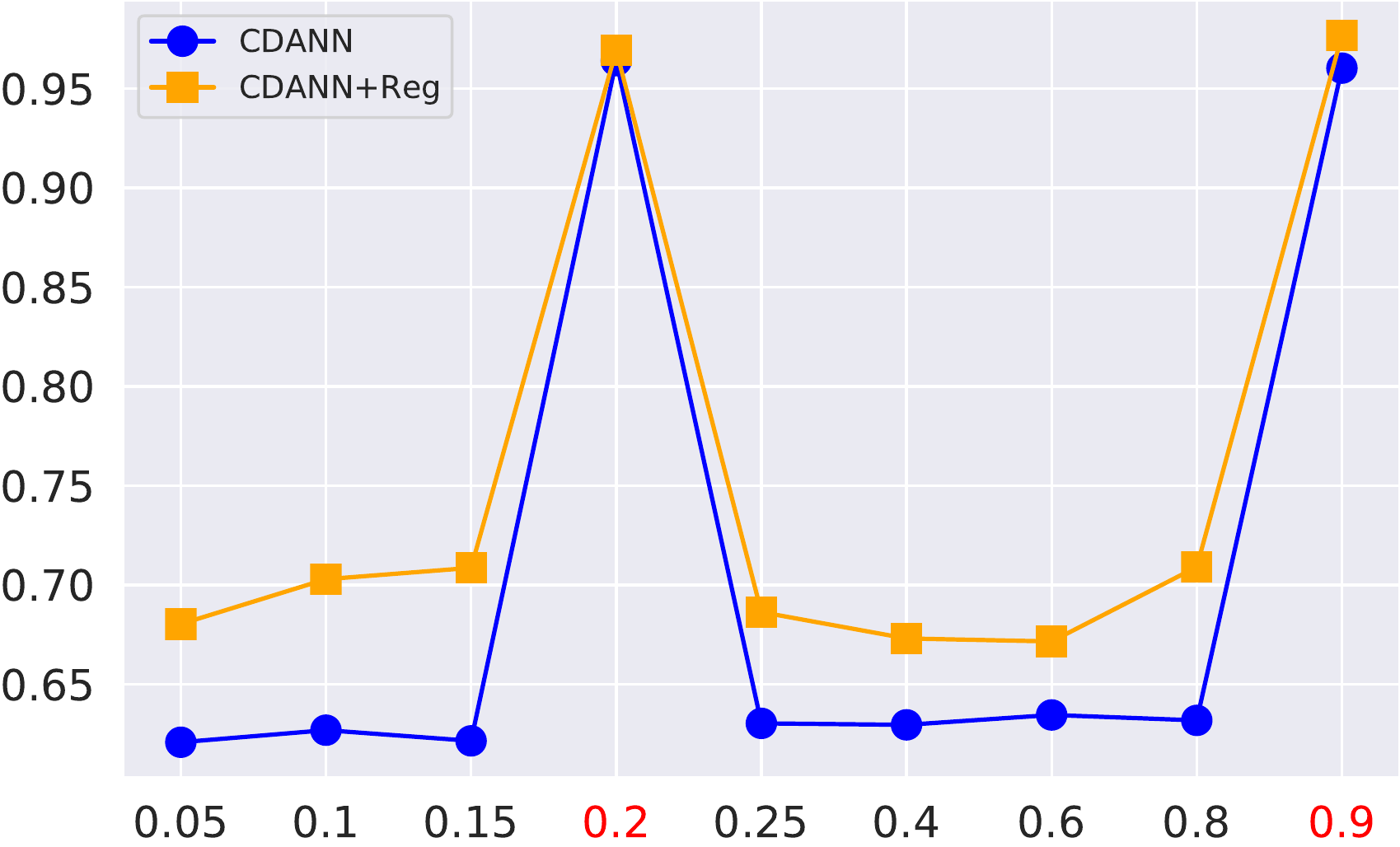}
    \caption{Generalization on different test environments. The observed environments are $P_{\calS}=\{\textcolor{red}{0.2},\textcolor{red}{0.9}\}$ with high prediction performance. However, in the generalization of other test environments, the regularization term consistently improves the prediction performance.}
    \label{fig:control_env}
\end{figure}

\paragraph{Generalization in controllable environment} In the ColorMNIST, we fix the observed environments as $P_{\calS}=\{0.2, 0.9\}$ and test on various environment with different $P_{\calT}=\{0.05,\dots,0.85\}$, shown in Fig.~\ref{fig:control_env}.  In the observed environments $P_{\calS}=\{0.2, 0.9\}$, both approaches achieve high prediction accuracy with $>95\%$. However, the generalization behaviors are quite different: adding an regularization term consistently improves the performance in out-of-distribution prediction through $3-5\%$.

\section{Conclusion}
In this paper, we analyzed the representation learning based domain generalization. Concretely, we highlight the importance of regularizing the representation function. Then we theoretically demonstrate the benefits of regularization, as the key role to control the prediction error in the unseen test environment. In practice, we evaluate the Jacobin matrix regularization on various invariance criteria and datasets, which suggests the benefits of regularization.

\section*{Appendix: Proof}

\textbf{Proof of Theorem 1} The prediction error on the test environment can be written as:
\begin{align*}
\mathrm{BER}_T(h,\Phi) & = \frac{1}{|\calY|}\sum_{y=1}^{\calY} \int_{z} T(z|Y=y)\calL(h(z),y)   \\
& \leq \frac{1}{|\calY|}\sum_{y=1}^{\calY} \left[ \E_{z \sim \calS^{\star}(z|Y=y)}\calL(h(z),y) + d_{\mathrm{TV}}(\calS^{\star}(z|Y=y)\|\calT(z|Y=y)) \right]
\end{align*}
Where $\calS^{\star}$ is the nearest source environment that is the most similar to the test environment (i.e.  in the raw feature space, $d_{\mathrm{TV}}(\calS^{\star}(x|Y=y)\|\calT(x|Y=y))\leq \epsilon$)
We have the following upper since the prediction loss in upper bounded by 1 and the property of TV distance (\cite{Polyanskiy2019}, Remark 3.1).

We first bound the first term, since $\calS^{\star}$ is unknown source during the training, then we can upper bound through all the sources, i.e, $\forall t\in\{1,\dots,T\}$, we have:
\[
\E_{z \sim \calS^{\star}(z|Y=y)}\calL(h(z),y) \leq \frac{1}{T}\sum_{t=1}^T \E_{z \sim \calS_{t}(z|Y=y)}\calL(h(z),y) + \frac{1}{T}\sum_{t=1}^T d_{\mathrm{TV}}(\calS^{\star}(z|Y=y)\|\calS_t(z|Y=y))
\]
The proof of the above inequality is analogous to the first inequality and derived by the property of TV distance. Concretely, we use the inequality $T$-times and then derive the average upper bound.

Since we adopt the feature conditional invariance criteria, then we have $d_{\mathrm{TV}}(\calS^{\star}(z|Y=y)\|\calS_t(z|Y=y))\leq \kappa$. This inequality holds since in training we have enforced small conditional invariance among all the sources. Then the term can be upper bounded by:
\[
\E_{z \sim \calS^{\star}(z|Y=y)}\calL(h(z),y) \leq \frac{1}{T}\sum_{t=1}^T \E_{z \sim \calS_{t}(z|Y=y)}\calL(h(z),y) + \kappa
\]

Then we upper bound the second term through introducing the strong data-processing inequality \cite{Polyanskiy2019}. The strong data-processing suggests a tighter bound of data-processing inequality. Specifically, it reveals the decay rate of information loss, characterized by the Dobrushin coefficient. 

\paragraph{Strong data-processing inequality} For distributions $P_0,P_1$ defined on $\calX$ and a channel $Q$ from space $\calX$ to space $\calZ$, define a marginal distribution $M_0(z) = \int Q(z|x)P_0(x)dx$. The channel $Q$ satisfies a strong data processing inequality with constant $\alpha\leq 1$ for the given $f$-divergence.
\begin{equation*}
    D_f(M_0\|M_1) \leq \alpha_f D_f(P_0\|P_1)
\end{equation*}
Where $\alpha$ is a constant defined with $\alpha_f(Q) = \sup_{P_0 \neq P_1} \frac{ D_f(M_0\|M_1)}{D_f(P_0\|P_1)}$.
For any convex $f$ divergence, we have:
\begin{equation*}
    \alpha_f(Q)\leq \alpha_{\text{TV}}(Q)
\end{equation*}
Where $\alpha_{\text{TV}}(Q)$ is the \emph{Dobrushin coefficient}, which is equivalent as:
\begin{equation*}
    \alpha_{\text{TV}}(Q) :=\sup_{x,x^{\prime}}d_{\text{TV}}(Q(\cdot|x)\|Q(\cdot|x^{\prime}))
\end{equation*}

In our problem, the embedding distribution $\Phi$ can be viewed as the information channel, and we denote distributions $P_0$ and $P_1$ as $\calS^{\star}(x|Y=y)$ and $\calT(x|Y=y)$. Then we have the conditional distribution defined on the latent space $\calS^{\star}(z|Y=y) = \int \Phi(z|x)\calS^{\star}(x|Y=y)dx$, $\calT(z|Y=y) = \int \Phi(z|x)\calT(x|Y=y)dx$
\[
d_{\text{TV}}(\calS^{\star}(z|y)\|\calT(z|y)) \leq \alpha_{\text{TV}}(\Phi) d_{\mathrm{TV}}(\calS^{\star}(z|Y=y)\|\calT(z|Y=y)) \leq \alpha_{\text{TV}}(\Phi)\epsilon
\]

Plugging in all the elements, we have the upper bound:
\[
\mathrm{BER}_T(h,\Phi) \leq \frac{1}{|\calY|}\sum_{y=1}^{\calY}(\frac{1}{T}\sum_{t=1}^T \E_{z \sim \calS_{t}(z|Y=y)}\calL(h(z),y) + \kappa + \alpha_{\text{TV}}(\Phi)\epsilon)
\]
Rearranging the results, we have:
\[ 
\text{BER}_{\calT}(h,\Phi) \leq \frac{1}{T} \sum_{t=1}^T \text{BER}_{\calS_t}(h,\Phi) + \kappa + \alpha_{\mathrm{TV}}(\Phi)\epsilon
\]

\noindent \textbf{Proof of Lemma 1} We first prove the relation with feature conditional invariance and marginal invariance.

\paragraph{Relation with marginal invariance} According to the definition, we have:
\begin{align*}
    \E_{z\sim\Omega^{\star}}|\calS_i(z)-\calS_j(z)| & =  \E_{z\sim\Omega^{\star}}|\sum_{y}\calS_i(y)\calS_i(z|y)-\sum_{y}\calS_j(y)\calS_j(z|y)|\\
    & = \frac{1}{|\calY|}\E_{z\sim\Omega^{\star}} |\sum_{y}(\calS_i(z|y)-\calS_j(z|y))| \\
    & \leq \frac{1}{|\calY|} \sum_{y} \E_{z\sim\Omega^{\star}}|\calS_i(z|y)-\calS_j(z|y)|\\
    & = \frac{1}{|\calY|} \sum_{y} d_{\text{TV}}(\calS_i(z|y)\|\calS_j(z|y)) \leq \kappa
\end{align*}

\paragraph{Relation with label conditional invariance} According to the definition, we have:
\begin{align*}
    \E_{z\sim\Omega^{\star}} |\calS_i(y|z)-\calS_j(y|z)| & = \int_{z\sim\Omega^{\star}}  |\frac{\calS_i(z|y)\calS_i(y)}{\sum_{y}\calS_i(y)\calS_i(z|y)} - \frac{\calS_j(z|y)\calS_j(y)}{\sum_{y}\calS_j(y)\calS_j(z|y)}|\\
    & = \int_{z\sim\Omega^{\star}} |\frac{\calS_i(z|y)}{\sum_{y}\calS_i(z|y)} - \frac{\calS_j(z|y)}{\sum_{y}\calS_j(z|y)}|\\
    & \leq \int_{z\sim\Omega^{\star}} |\frac{\calS_i(z|y)}{\sum_{y} \calS_i(z|y)} - \frac{\calS_i(z|y)}{\sum_{y} \calS_j(z|y)}| + |\frac{\calS_i(z|y)}{\sum_{y} \calS_j(z|y)} - \frac{\calS_j(z|y)}{\sum_{y} \calS_j(z|y)} | \\
\end{align*}
We start to upper bound this two terms. For the first term, we have:
\begin{align*}
    \int_{z\sim\Omega^{\star}} |\frac{\calS_i(z|y)}{\sum_{y} \calS_i(z|y)} - \frac{\calS_i(z|y)}{\sum_{y} \calS_j(z|y)}| & = \int_{z\sim\Omega^{\star}} \calS_i(z|y)\frac{|\sum_y [\calS_j (z|y) - \calS_i (z|y)]|}{[\sum_{y}\calS_i(z|y)][\sum_{y}\calS_j(z|y)]}\\
    & = \int_{z\sim\Omega^{\star}} \frac{\calS_i(z|y)}{\sum_{y}\calS_i(z|y)} \frac{\sum_{y}|\calS_j (z|y) - \calS_i (z|y)|}{\sum_{y}\calS_j(z|y)}\\
    & \leq \int_{z\sim\Omega^{\star}} \frac{\sum_{y}|\calS_j (z|y) - \calS_i (z|y)|}{\sum_{y}\calS_j(z|y)} \\
    & \leq C_1 \sum_{y} \int_{z} |\calS_j (z|y) - \calS_i (z|y)| \\
    & \leq C_1 |\calY| \kappa
\end{align*}
Where $C_1 = \frac{1}{\inf_{z\in\Omega^{\star}}\sum_{y} \calS_j(z|y)}$ and we can verify $C_1>0$ since $\Omega^{\star}$ is the intersection region with non-zero measure. 

Then we bound the second term:

\begin{align*}
    \int_{z} |\frac{\calS_i(z|y)}{\sum_{y} \calS_j(z|y)} - \frac{\calS_j(z|y)}{\sum_{y} \calS_j(z|y)}| \leq
   \frac{1}{\inf_{z\in\Omega^{\star}}\sum_{y} \calS_j(z|y)} \int_{z}|\calS_i(z|y)-\calS_j(z|y)| = C_1 \kappa
\end{align*}

Combining all the results, we have:
\[
\E_{z\sim\Omega^{\star}} |\calS_i(y|z)-\calS_j(y|z)| \leq C_1(1+|\calY|)\kappa = C^{+}\kappa
\]
Where $C^{+}= C_1(1+|\calY|)$ is a positive constant. 

\section*{Proof of Lemma 2}
Since we approximate the $\Phi$ as a multi-dimensional Gaussian distribution, then the Dobrushin Coefficient can be computed as:
\begin{align*}
\sup_{x,x^{\prime}}~d_{\mathrm{TV}}(\mathcal{N}(\phi(x),\sigma^2 \mathbf{I}_{d})\| \mathcal{N}(\phi(x^{\prime}),\sigma^2 \mathbf{I}_{d}))    
\end{align*}
Since the TV distance of multi-dimensional Gaussian is infeasible to compute, then according to \cite{devroye2018total}, the upper bound of TV distance between two high-dimensional Gaussian distribution is:
\[
d_{\mathrm{TV}}(\mathcal{N}(\phi(x),\sigma^2 \mathbf{I}_{d})\| \mathcal{N}(\phi(x^{\prime}),\sigma^2 \mathbf{I}_{d}))  \leq \sqrt{2} d_{H}(\mathcal{N}(\phi(x),\sigma^2 \mathbf{I}_{d})\| \mathcal{N}(\phi(x^{\prime}),\sigma^2 \mathbf{I}_{d}))
\]
Where $d_{H}$ is the Hellinger distance, which has the closed form of between two Gaussian distributions with 
\begin{equation*}
    d_{H}(\mathcal{N}(\phi(x),\sigma^2 \mathbf{I}_{d}), \mathcal{N}(\phi(x^{\prime}),\sigma^2 \mathbf{I}_{d})) = \left(1- \exp(-\frac{1}{8\sigma^2 d} [\phi(x)-\phi(x^{\prime})]^{T}  [\phi(x)-\phi(x^{\prime})])\right)^{1/2}
\end{equation*}
Then the TV distance can be upper bounded as:
\[
\alpha_{\mathrm{TV}}(\Phi) \leq \sup_{x,x^{\prime}\in\calX} \sqrt{2}\left(1- \exp(-\frac{1}{8d \sigma^2} \|\phi(x)-\phi(x^{\prime})\|^2 )\right)^{1/2}
\]
We assume $\phi$ is $L_{\phi}$ Lipschitz such that w.r.t. $x$,
\[
\|\phi(x)-\phi(x^{\prime})\| \leq L_{\phi} \|x-x^{\prime} \|_2
\]
and the $d_{\max} = \sup_{x,x^{\prime}}\|x-x^{\prime} \|_2$. Then we have:
\[
\alpha_{\mathrm{TV}}(\Phi) \leq \sqrt{2}\left(1- \exp(-\frac{d^2_{\max}}{8d \sigma^2} L^2_{\phi} )\right)^{1/2} 
\]

\section*{Relation with Data-Augmentation}
In this part, we will demonstrate a simple proof to show the role of data-augmentation, which also aims at regularizing the representation function.

We suppose a differentiable embedding function $\phi:\calX\to\calZ$ and we suppose the loss function as logistic loss $\calL(\hat{y},y) = \log(1+\exp(-\hat{y}y))$ and the predictor as a linear function $w$, binary classification with balanced label distribution. Then the objective function can be written as:
\[ \calG(w) = \E_{\tilde{x}}~\calL(w^{T}\phi(\tilde{x}),y)\]
Where $\tilde{x}=\text{INP}(x_1,\dots,x_T), x_1\sim\calS_1(x|Y=y), \dots, x_T\sim\calS_T(x|Y=y)$ is any interpolation function of samples from multiple environments. 
We also suppose the data-augmentation aims at improving the local property of the representation $\phi$. Then by using first-order Taylor expansion at local representation $\phi_0$, we have:
\[
\calG_1(w) = \calL(w^{T}\phi_0, y) + \E_{\tilde{x}} (\phi_0 - \phi(\tilde{x})) \calL^{\prime}(w^{T}\phi_0,y)
\]
If we take $\phi_0(x) = \E_{\tilde{x}} [\phi(\tilde{x})]$, then the second term vanish, then the first order approximation can be expressed as:
\[
\calG_1(w) = \calL(w^{T} \E_{\tilde{x}}[\phi(x)], y) 
\]
Then we compute the second-order approximation at point $\phi_0$, then we have
\[
\calG_2(w) = \frac{1}{2} \E_{\tilde{x}} [(w^{T} (\phi(\tilde{x})-\E_{\tilde{x}} [\phi(\tilde{x})]))^2 \calL^{\prime\prime}(w^{T}\E_{\tilde{x}} [\phi(\tilde{x})],y)
\]
We can further compute that if $\calL$ is logistic loss, the second-derivative is independent of label $y$ and the second derivative is bounded by $1$. Then we have
\[
\calG_2(w) \leq \frac{1}{2} \text{Var}_{\tilde{x}} (w^{T}\phi(\tilde{x}))^2
\]

\paragraph{Relation with regularization term} If the embedding function is $L_{\phi}$ Lipschitz then the function $w^{T}\phi(\tilde{x})$ is also $L_{\phi}\|w\|_2$-Lipschitz through:
\[
|w^{T}\phi(\tilde{x}_1) - w^{T}\phi(\tilde{x}_2)| \leq \|w\|_2 \|\phi(\tilde{x}_1)-\phi(\tilde{x}_2)\|_2 \leq L_{\phi}\|w\|_2 \|\tilde{x}_1-\tilde{x}_2\|
\]
Then we have the upper bound of $\calG_2 \leq L^2_{\phi} \frac{\|w\|^2_2}{4}\text{Var}(\tilde{x})$. Therefore, the minimize the loss on the augmented-data set can be viewed as an implicit optimization to enforce a small prediction variance, where the Lipschitz representation function $\phi$ is one sufficient condition to realize it.

\begin{proof}
We can compute the second derivative of $\calL(\hat{y},y) = \log(1+\exp(-\hat{y}y))$ w.r.t. $\hat{y}$:
\[ \frac{\partial^2 \calL(\hat{y},y)}{\partial \hat{y}^2} = \frac{y^2\exp(y\hat{y})}{(1+\exp(y\hat{y}))^2}\]
Since $y$ is binary with possible values $y=\{-1,+1\}$, then we have $y^2=1$, the second-derivative is independent of $y$ with $\frac{\partial^2 \calL(\hat{y},y=1)}{\partial \hat{y}^2} = \frac{\partial^2 \calL(\hat{y},y=-1)}{\partial \hat{y}^2} = \frac{\exp(\hat{y})}{(1+\exp(\hat{y}))^2} \leq 1 $
\end{proof}

\section*{Appendix: The Network Structure}
\begin{figure}[h]
    \centering
    \begin{enumerate}
     \item Feature extractor: with 3 convolution layers. 
    
    'layer1': 'conv': [3, 3, 64], 'relu': [], 'maxpool': [2, 2, 0],
    
    'layer2': 'conv': [3, 3, 128], 'relu': [], 'maxpool': [2, 2, 0],
    
    'layer3': 'conv': [3, 3, 256], 'relu': [], 'maxpool': [2, 2, 0],
    
    \item Task prediction: with 3 fully connected layers.
    
    'layer1': 'fc': [*, 512], 'act\_fn': 'relu',
    
    'layer2': 'fc': [512, 100], 'act\_fn': 'relu',
    
    'layer3': 'fc': [100, 2],
    
    \item Domain Discriminator:  with 2 fully connected layers.
    
    \emph{reverse\_gradient}()
    
    'layer1': 'fc': [*, 256], 'act\_fn': 'relu',

    'layer2': 'fc': [256, 2], 
\end{enumerate}
    \caption{Neural Network Structure in the digits recognition}
    \label{network_str}
\end{figure}

\begin{figure}[h]
    \centering
    \begin{enumerate}
     \item Feature extractor: ResNet18,
    
    \item Task prediction: with 3 fully connected layers.
    
    'layer1': 'fc': [*, 256], 'batch\_normalization', 'act\_fn': 'Leaky\_relu',
    
    'layer2': 'fc': [256, 256], 'batch\_normalization', 'act\_fn': 'Leaky\_relu',
    
    'layer3': 'fc': [256, class\_number], 
    
    \item Domain Discriminator:  with 3 fully connected layers.
    
    \emph{reverse\_gradient}()
    
    'layer1': 'fc': [*, 256], 'batch\_normalization', 'act\_fn': 'Leaky\_relu',
    
    'layer2': 'fc': [256, 256], 'batch\_normalization', 'act\_fn': 'Leaky\_relu', 
    
    'layer3': 'fc': [256, class\_number], 'Sigmoid', 
\end{enumerate}
    \caption{Neural Network Structure in the PACS/Office-Home}
    \label{network_str_office}
\end{figure}

\bibliographystyle{spmpsci}      
\bibliography{ref.bib}   

\begin{thebibliography}{10}
\providecommand{\url}[1]{{#1}}
\providecommand{\urlprefix}{URL }
\expandafter\ifx\csname urlstyle\endcsname\relax
  \providecommand{\doi}[1]{DOI~\discretionary{}{}{}#1}\else
  \providecommand{\doi}{DOI~\discretionary{}{}{}\begingroup
  \urlstyle{rm}\Url}\fi

\bibitem{achille2018emergence}
Achille, A., Soatto, S.: Emergence of invariance and disentanglement in deep
  representations.
\newblock The Journal of Machine Learning Research \textbf{19}(1), 1947--1980
  (2018)

\bibitem{albuquerque2019generalizing}
Albuquerque, I., Monteiro, J., Darvishi, M., Falk, T.H., Mitliagkas, I.:
  Generalizing to unseen domains via distribution matching.
\newblock arXiv preprint arXiv:1911.00804  (2019)

\bibitem{arjovsky2019invariant}
Arjovsky, M., Bottou, L., Gulrajani, I., Lopez-Paz, D.: Invariant risk
  minimization.
\newblock arXiv preprint arXiv:1907.02893  (2019)

\bibitem{baxter2000model}
Baxter, J.: A model of inductive bias learning.
\newblock Journal of artificial intelligence research \textbf{12}, 149--198
  (2000)

\bibitem{ben2010theory}
Ben-David, S., Blitzer, J., Crammer, K., Kulesza, A., Pereira, F., Vaughan,
  J.W.: A theory of learning from different domains.
\newblock Machine learning \textbf{79}(1), 151--175 (2010)

\bibitem{buhlmann2020invariance}
B{\"u}hlmann, P., et~al.: Invariance, causality and robustness.
\newblock Statistical Science \textbf{35}(3), 404--426 (2020)

\bibitem{devroye2018total}
Devroye, L., Mehrabian, A., Reddad, T.: The total variation distance between
  high-dimensional gaussians.
\newblock arXiv preprint arXiv:1810.08693  (2018)

\bibitem{ganin2016domain}
Ganin, Y., Ustinova, E., Ajakan, H., Germain, P., Larochelle, H., Laviolette,
  F., Marchand, M., Lempitsky, V.: Domain-adversarial training of neural
  networks.
\newblock The Journal of Machine Learning Research \textbf{17}(1), 2096--2030
  (2016)

\bibitem{goodfellow2014explaining}
Goodfellow, I.J., Shlens, J., Szegedy, C.: Explaining and harnessing
  adversarial examples.
\newblock arXiv preprint arXiv:1412.6572  (2014)

\bibitem{gulrajani2021in}
Gulrajani, I., Lopez-Paz, D.: In search of lost domain generalization.
\newblock In: International Conference on Learning Representations (2021).
\newblock \urlprefix\url{https://openreview.net/forum?id=lQdXeXDoWtI}

\bibitem{ilse2019diva}
Ilse, M., Tomczak, J.M., Louizos, C., Welling, M.: Diva: Domain invariant
  variational autoencoders.
\newblock arXiv preprint arXiv:1905.10427  (2019)

\bibitem{kamath2021does}
Kamath, P., Tangella, A., Sutherland, D.J., Srebro, N.: Does invariant risk
  minimization capture invariance?
\newblock arXiv preprint arXiv:2101.01134  (2021)

\bibitem{li2017deeper}
Li, D., Yang, Y., Song, Y.Z., Hospedales, T.M.: Deeper, broader and artier
  domain generalization.
\newblock In: Proceedings of the IEEE international conference on computer
  vision, pp. 5542--5550 (2017)

\bibitem{li2018learning}
Li, D., Yang, Y., Song, Y.Z., Hospedales, T.M.: Learning to generalize:
  Meta-learning for domain generalization.
\newblock In: Thirty-Second AAAI Conference on Artificial Intelligence (2018)

\bibitem{li2018domain}
Li, Y., Gong, M., Tian, X., Liu, T., Tao, D.: Domain generalization via
  conditional invariant representations.
\newblock In: Proceedings of the AAAI Conference on Artificial Intelligence,
  vol.~32 (2018)

\bibitem{li2019feature}
Li, Y., Yang, Y., Zhou, W., Hospedales, T.M.: Feature-critic networks for
  heterogeneous domain generalization.
\newblock arXiv preprint arXiv:1901.11448  (2019)

\bibitem{lu2021nonlinear}
Lu, C., Wu, Y., Hern{\'a}ndez-Lobato, J.M., Sch{\"o}lkopf, B.: Nonlinear
  invariant risk minimization: A causal approach.
\newblock arXiv preprint arXiv:2102.12353  (2021)

\bibitem{dg_mmld}
Matsuura, T., Harada, T.: Domain generalization using a mixture of multiple
  latent domains.
\newblock In: AAAI (2020)

\bibitem{mirza2014conditional}
Mirza, M., Osindero, S.: Conditional generative adversarial nets.
\newblock arXiv preprint arXiv:1411.1784  (2014)

\bibitem{miyato2018spectral}
Miyato, T., Kataoka, T., Koyama, M., Yoshida, Y.: Spectral normalization for
  generative adversarial networks.
\newblock arXiv preprint arXiv:1802.05957  (2018)

\bibitem{muller2020learning}
M{\"u}ller, J., Schmier, R., Ardizzone, L., Rother, C., K{\"o}the, U.: Learning
  robust models using the principle of independent causal mechanisms.
\newblock arXiv preprint arXiv:2010.07167  (2020)

\bibitem{Polyanskiy2019}
Polyanskiy, Y., Wu, Y.: Lecture notes on information theory (2019)

\bibitem{roberts2021sgd}
Roberts, D.A.: Sgd implicitly regularizes generalization error.
\newblock arXiv preprint arXiv:2104.04874  (2021)

\bibitem{sicilia2021domain}
Sicilia, A., Zhao, X., Hwang, S.J.: Domain adversarial neural networks for
  domain generalization: When it works and how to improve.
\newblock arXiv preprint arXiv:2102.03924  (2021)

\bibitem{sugiyama2007covariate}
Sugiyama, M., Krauledat, M., M{\"u}ller, K.R.: Covariate shift adaptation by
  importance weighted cross validation.
\newblock Journal of Machine Learning Research \textbf{8}(5) (2007)

\bibitem{venkateswara2017deep}
Venkateswara, H., Eusebio, J., Chakraborty, S., Panchanathan, S.: Deep hashing
  network for unsupervised domain adaptation.
\newblock In: Proceedings of the IEEE conference on computer vision and pattern
  recognition, pp. 5018--5027 (2017)

\bibitem{volpi2018generalizing}
Volpi, R., Namkoong, H., Sener, O., Duchi, J., Murino, V., Savarese, S.:
  Generalizing to unseen domains via adversarial data augmentation.
\newblock arXiv preprint arXiv:1805.12018  (2018)

\bibitem{wang2020domainmix}
Wang, W., Liao, S., Zhao, F., Kang, C., Shao, L.: Domainmix: Learning
  generalizable person re-identification without human annotations.
\newblock arXiv preprint arXiv:2011.11953  (2020)

\bibitem{zhang2013domain}
Zhang, K., Sch{\"o}lkopf, B., Muandet, K., Wang, Z.: Domain adaptation under
  target and conditional shift.
\newblock In: International Conference on Machine Learning, pp. 819--827. PMLR
  (2013)

\bibitem{zhao2020domain}
Zhao, S., Gong, M., Liu, T., Fu, H., Tao, D.: Domain generalization via entropy
  regularization.
\newblock Advances in Neural Information Processing Systems \textbf{33} (2020)

\bibitem{zhou2020learning}
Zhou, K., Yang, Y., Hospedales, T., Xiang, T.: Learning to generate novel
  domains for domain generalization.
\newblock In: European Conference on Computer Vision, pp. 561--578. Springer
  (2020)

\bibitem{zhou2021domain}
Zhou, K., Yang, Y., Qiao, Y., Xiang, T.: Domain generalization with mixstyle.
\newblock arXiv preprint arXiv:2104.02008  (2021)

\end{thebibliography}

\end{document}